\documentclass[conference]{IEEEtran}
\IEEEoverridecommandlockouts
\usepackage{cite}
\usepackage{url}
\usepackage{amsmath,amssymb,amsfonts}
\usepackage{algorithmic}
\usepackage{graphicx}
\usepackage{textcomp}
\usepackage{xcolor}
\usepackage{multirow} 
\usepackage{adjustbox}
\usepackage{array}
\usepackage{float}
\usepackage{subcaption} 
\usepackage{booktabs}
\def\BibTeX{{\rm B\kern-.05em{\sc i\kern-.025em b}\kern-.08em
    T\kern-.1667em\lower.7ex\hbox{E}\kern-.125emX}}
\begin{document}

\title{Differentiable Biomechanics for Markerless Motion Capture in Upper Limb Stroke Rehabilitation: A Comparison with Optical Motion Capture \\
\thanks{part of this work is funded by Digitalisierungsinitiative der Zürcher Hochschulen (DIZH)}
}

\author{\IEEEauthorblockN{1\textsuperscript{st*} Tim Unger}
\IEEEauthorblockA{\textit{DART Lab, LLUI, Vitznau} \\
\textit{RELab, ETH, Zurich}\\
Switzerland \\
tim.unger@llui.org} 
\and
\IEEEauthorblockN{2\textsuperscript{nd*} Arash Sal Moslehian}
\IEEEauthorblockA{\textit{Neuro-X Institute} \\
\textit{EPFL}\\
Lausanne, Switzerland \\
arash.salmoslehian@epfl.ch}
\and
\IEEEauthorblockN{3\textsuperscript{rd*} J.D. Peiffer}
\IEEEauthorblockA{\textit{Dept of Biomedical Engineering} \\
\textit{Northwestern University}\\
Chicago, IL \\
jpeiffer@sralab.org}
\and
\IEEEauthorblockN{4\textsuperscript{th} Johann Ullrich}
\IEEEauthorblockA{\textit{DART Lab, LLUI, Vitznau} \\
\textit{RELab, ETH, Zurich}\\
Switzerland \\
jullrich@ethz.ch}
\and
\IEEEauthorblockN{5\textsuperscript{th} Roger Gassert}
\IEEEauthorblockA{\textit{Rehabilitation Engineering Laboratory (RELab)} \\
\textit{ETH}\\
Zurich, Switzerland \\
roger.gassert@hest.ethz.ch}
\and
\IEEEauthorblockN{6\textsuperscript{th} Olivier Lambercy}
\IEEEauthorblockA{\textit{RELab} \\
\textit{ETH}\\
Zurich, Switzerland \\
olivier.lambercy@hest.ethz.ch}
\and
\IEEEauthorblockN{7\textsuperscript{th}\textsuperscript{$\dagger$} R. James Cotton}
\IEEEauthorblockA{\textit{Shirley Ryan AbilityLab} \\
\textit{Dept of PM\&R, Northwestern University}\\
Chicago, IL \\
rcotton@sralab.org}
\and
\IEEEauthorblockN{8\textsuperscript{th}\textsuperscript{$\dagger$} Chris Awai Easthope}
\IEEEauthorblockA{\textit{Data Analytics \& Rehabilitation Technology (DART Lab)} \\
\textit{Lake Lucerne Institute (LLUI)}\\
Vitznau, Switzerland \\
chris.awai@llui.org}
\and
\IEEEauthorblockN{* co-first author \\ $\dagger$ co-senior author}
}

\maketitle

\begin{abstract}
Marker-based Optical Motion Capture (OMC) paired with biomechanical modeling is currently considered the most precise and accurate method for measuring human movement kinematics. However, combining differentiable biomechanical modeling with Markerless Motion Capture (MMC) offers a promising approach to motion capture in clinical settings, requiring only minimal equipment, such as synchronized webcams, and minimal effort for data collection. 

This study compares key kinematic outcomes from biomechanically modeled MMC and OMC data in 15 stroke patients performing the drinking task, a functional task recommended for assessing upper limb movement quality. 

We observed a high level of agreement in kinematic trajectories between MMC and OMC, as indicated by high correlations (median $r > 0.95$ for the majority of kinematic trajectories) and median $\text{RMSE}$ values ranging from $2^\circ$–$5^\circ$ for joint angles, $0.04 \, \text{m/s}$ for end-effector velocity, and $6 \, \text{mm}$ for trunk displacement. Trial-to-trial biases between OMC and MMC were consistent within participant sessions, with interquartile ranges of bias around $1-3^\circ$ for joint angles, $0.01m/s$ in end-effector velocity, and approximately $3 \, \text{mm}$ for trunk displacement. 

Our findings indicate that our MMC for arm tracking is approaching the accuracy of marker-based methods, supporting its potential for use in clinical settings. MMC could provide valuable insights into movement rehabilitation in stroke patients, potentially enhancing the effectiveness of rehabilitation strategies. 
\end{abstract}
\begin{IEEEkeywords}
markerless motion capture, optical motion capture, arm biomechanics, stroke, drinking task
\end{IEEEkeywords}


\section{Introduction}
The increasing number of patients affected by stroke and related neurological conditions has created an urgent need to improve the effectiveness of rehabilitation strategies \cite{kwakkel_standardized_2017}. Over 70\% of stroke survivors suffer from acute upper limb impairment \cite{lawrence_estimates_2001}, necessitating accurate and frequent assessments of upper limb movement quality to customize interventions and monitor rehabilitation progress \cite{kwakkel_standardized_2019}. Ideally, recovery in stroke patients involves neuroplasticity, where motor functions are restored by reorganizing neural pathways \cite{su_enhancing_2020}. Compensatory strategies, such as relying on unaffected muscles or limbs, provide short-term functional benefits but can hinder neuroplastic recovery by reinforcing maladaptive patterns \cite{alia_neuroplastic_2017}. As a result, after controlling for altered movement due to pain, deviation in movement quality can serve as a biomarker for limited neuroplasticity and reduced genuine recovery \cite{solnik_movement_2020, Levin_2009}.

Consensus guidelines from European stroke experts recommend incorporating an instrumented drinking task to assess upper limb movement quality in stroke patients \cite{kwakkel_standardized_2019}. The movement quality measures for this task have been established using optical motion capture (OMC) in several studies \cite{ murphy_kinematic_2018, frykberg_how_2021,murphy_movement_2012}. Although OMC does not directly measure bone movement, it remains widely regarded as the gold standard for biomechanical measurement \cite{disidoro_moving_2023}. However, the high cost and logistical complexity of OMC limit its accessibility in routine clinical practice. However, recent advances in human pose estimation with computer vision now enable markerless motion capture (MMC) using multiple calibrated cameras, which may offer a feasible alternative to OMC \cite{lam_systematic_2023}.

Traditional biomechanical modeling with markers follows a two-stage process: first, 3D marker trajectories are captured and a stationary period is used to scale the biomechanical model; second, the marker trajectories are fit to that scaled biomechanical model with virtual markers, minimizing errors between observed and model marker. This process assumes fixed marker trajectories on known body landmarks, preventing measurement uncertainties from propagating through the model and thus requiring highly accurate marker placement and marker trajectories. Both OMC and MMC face challenges in meeting this requirement due to inconsistent marker placement \cite{uchida_conclusion_2022} and soft tissue artifacts \cite{fiorentino_soft_2017} in OMC, and keypoint detection uncertainty in MMC \cite{ruescas-nicolau_deep_2024}.

To address the limitations of traditional modeling, specifically in MMC, Cotton et al. proposed a differentiable end-to-end optimization algorithm allowing for simultaneous optimization of movement trajectories and model scaling parameters \cite{cotton_optimizing_2023, cotton_differentiable_2024} (hereafter referred to as end-to-end MMC). While this method is accurate based on low reprojection errors and spatiotemporal gait parameters \cite{firouzabadi_biomechanical_2024,cotton_optimizing_2023}, it has not yet been validated against OMC-based kinematic data.

In this study, we conduct the first comparative analysis between OMC and this novel end-to-end MMC approach. We evaluate the drinking task movement quality measures in stroke patients and compare the underlying kinematic trajectories used to derive these measures. Our hypothesis is that the agreement between the two systems will be high across key kinematic trajectories, such as joint angles, end-effector velocities, and trunk displacement as well as in the established movement quality measures for the drinking task. We test this hypothesis using comparative plots for kinematic trajectories and movement quality measures as well as several comparative metrics such RMSE, correlation, and bias to assess the agreement of movement quality measures and underlying kinematic trajectories produced by each system\footnote{The code for the analysis is available at \url{https://github.com/DART-Lab-LLUI/diff-biomech-mmc-icorr-2025}}.

\section{Methods}
\subsection{Participants}
The study adhered to the ethical guidelines set forth by the local ethics committee (BASEC-No: 2022-00491).  
Stroke survivors with mild and moderate upper limb impairment were recruited from the University Hospital Zurich Stroke Registry and the cereneo
clinic. Eligible participants were invited for a single-session
measurement. Inclusion criteria mandated that participants be at least 18 years
old, capable of providing informed consent, and have a confirmed
diagnosis of stroke. Additionally, participants were required to have
at least partial ability to perform a reaching movement, and to grasp
a cup unassisted using a cylindrical grip with the affected hand.
Exclusion criteria consisted of pre-existing upper limb deﬁcits,
such as orthopedic impairments, and other neurological conditions.

\subsection{Data Acquisition and Preprocessing}
\begin{figure}[htbp]
    \centering
    \includegraphics[width= \columnwidth]{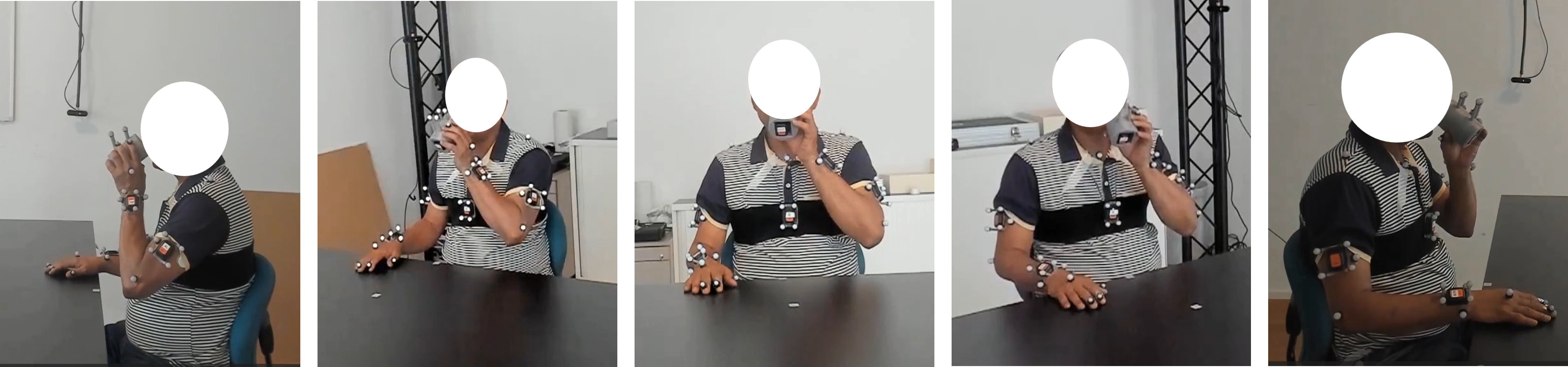}
    \caption{Camera perspectives of the five webcams used for MMC.}
    \label{fig:CameraPerspectives}
\end{figure}

Participants performed 40 repetitions of the standardized drinking task \cite{murphy_kinematic_2018} with each the affected and unaffected arm while being recorded with OMC as well as five synchronized consumer-grade webcams for MMC. The OMC system operated at a sampling frequency of 100Hz, using technologies from Qualisys and Optitrack across different measurement setups. We applied the base marker set used by Alt Murphy et al. \cite{murphy_kinematic_2018} for the drinking task and extended it with further markers for better representation of the movement. OMC data was manually gap-filled and labeled. The five synchronized Logitech Brio webcams recorded the participants at 1080p and 60Hz. The five cameras were evenly distributed in a half circle around the participant with a spacing of roughly 45° (see Fig. \ref{fig:CameraPerspectives}). 
The videos were segmented into trials based on the OMC master trigger signal, resulting in a synchronized dataset composed of multiple trials, each representing a single repetition of the standardized drinking task.   

\subsection{Data Processing and Inverse Kinematics}
\subsubsection{Two-Stage OMC}
The measured 3D markers from OMC during the initial static period of participants sitting at the table were used to scale a biomechancical model in OpenSim. The OpenSim model is based on the Rajagopal 2016 model \cite{rajagopal_full-body_2016}, with modified ranges of motion for pronation and supination to better match the natural range of motion, as the original model constrained these motions. As the legs were stationary and occluded by the table during the drinking task, we removed all markers below the hip, effectively locking all leg degrees of freedom in the standard pose.
Afterwards, inverse kinematics were calculated using the IK tool in OpenSim with the scaled model and the measured 3D marker trajectories from OMC. All calculations were performed using Python scripts, incorporating the OpenSim API to allow batch processing. Resulting kinematic trajectories of OMC were interpolated from 100Hz to 60Hz to compare directly with MMC.

\subsubsection{End-to-End MMC}
Previous work has shown that jointly solving for body scale parameters and joint angles in an end-to-end differentiable manner is an effective way to learn biomechanically aware trajectories \cite{cotton_differentiable_2024}. To apply this approach to the data collected here, we first used AniposeLib \cite{karashchuk_aniposelib_2024} to perform extrinsic and intrinsic calibration of cameras using a calibration recording with an A1-sized ChArUco board, required for triangulation of 2D keypoints. Using PosePipe \cite{cotton_posepipe_2022}, we first extracted 87 2D keypoints from each webcam video using MeTrabs-ACAE \cite{sarandi_learning_2022} and then performed automatic person association using EasyMocap \cite{noauthor_easymocap_2021}. For videos where automatic person association failed, we manually performed person association. 

Finally, we reconstructed joint kinematics following \cite{cotton_differentiable_2024}. In brief, this parameterizes the movement trajectory of a recording as an implicit function $f_\phi: t \rightarrow \theta$ mapping from time to joint angles where $f_{\phi}$ is implemented as a multi-layer perceptron. These parameters $\phi$ along with body scaling and marker offset parameters are jointly learned through optimization. At each optimization step, the joint angle output from the implicit function is passed through a biomechanical forward kinematic function in MuJoCo using the GPU accelerated Mjx version \cite{todorov_mujoco_2012,caggiano_myosuite_2022} and 3D pose marker locations are projected into each camera frame using known camera extrinsics/intrinsics. This approach also employs a bilevel optimization which jointly learns the body scale per participant with the inverse kinematics over multiple trials \cite{werling_rapid_2022}. Since each of our participants presented here had approximately 80 trials, not all the trials could fit into our NVIDIA RTX 3080 GPU with 10GB of VRAM; thus, we performed bilevel optimization on a subset of each participant's trials, splitting them into eight batches. 
We used a modified version of the biomechanical model presented by Hamner et al. \cite{hamner_muscle_2010, al-hafez_locomujoco_2023} compatible with Mujoco. We found no major differences in relevant model degrees of freedom between the models used for OMC (OpenSim) and MMC (MuJoCo).
\begin{figure}[b]
    \centering
    \includegraphics[width=\columnwidth]{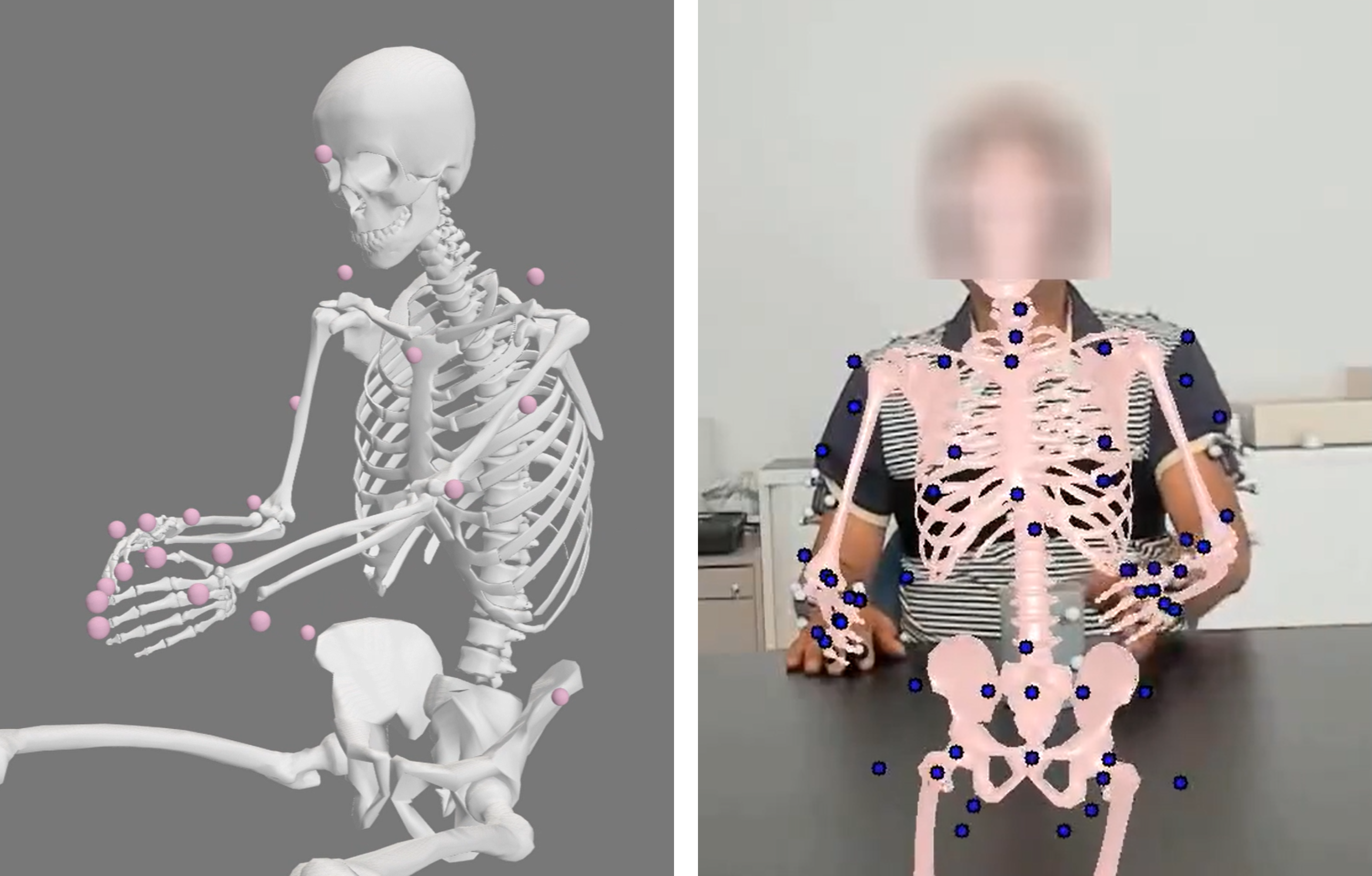}
    \caption{Visualization of biomechnical modelling. left: OMC data with Opensim; right: end-to-end MMC with Mujoco.}
    \label{fig:Visualization}
\end{figure}

\subsection{System Comparison}

\subsubsection{Kinematic Trajectories}
We included the essential kinematic trajectories (kinematic time series data, such as joint angles, displacements and velocity profiles)  for deriving movement quality measures in stroke patients performing the drinking task according to \cite{murphy_kinematic_2018}. Specifically, these kinematic trajectories included: Shoulder Flexion, Shoulder Abduction, Elbow Flexion, Elbow Angular Velocity, Trunk Displacement, and End-Effector Velocity.

Joint angles were extracted from the inverse kinematics of biomechanical models in both OMC, using OpenSim and MMC, using MuJoCo. The end-effector velocity for OMC was computed by OpenSim based on the velocity of the center of geometry of the hand marker, while for MMC we differentiated the position of the hand keypoint. Trunk displacement was determined from the position of the trunk marker (sternum keypoint for MMC) after model fitting.

For each trial and kinematic trajectory, we calculated the static offset between the two systems as the difference in the means of MMC and OMC trajectory. This bias (offset) was then added to the OMC signal. The static bias for each kinematic trajectory was recorded to evaluate the consistency of the offsets between the two methods across trials and participants. All further analyses of noise were conducted with bias-removed kinematic trajectories.

To account for potential time offsets due to imperfect synchronization between the systems, we determined the optimal time lag between the two signals by shifting one kinematic trajectory over the other, up to 0.25 seconds. At each shift, we calculated the RMSE and identified the time lag that minimized this error. The resulting time lag was recorded as a measure of synchronization accuracy. Finally, we calculated the RMSE and Pearson correlation between the adjusted kinematic trajectories from OMC and MMC for each trial to assess the agreement between the two systems.

\subsubsection{Movement Quality Measures}

Movement quality measures for the drinking task, based on what was defined by Alt Murphy et al. \cite{murphy_kinematic_2018}, were derived from the kinematic trajectories of both MMC and OMC.
These include: total movement time, number of movement units (smoothness), peak velocity (PV) in reaching, peak elbow angular velocity in reaching, time to peak velocity in reaching, time to first peak velocity in reaching, max elbow extension in reaching, max shoulder abduction angle in drinking / reaching, max trunk displacement during task, max shoulder flexion angle in reaching / drinking, interjoint coordination (Pearson correlation between elbow extension and shoulder flexion during reaching).
Drinking task phase classification (Reaching, Forward, Drinking, Back, Returning, and Rest) for both systems was based on the end-effector velocity calculated from the OMC markers on the wrist and cup, following the approach in \cite{murphy_kinematic_2018}. This eliminated any influence of difference in phase classification, thereby isolating the focus on kinematic data alone. Therefore, time to peak velocity and time to first peak velocity during reaching (in \%) are not reported, as the phase duration is identical for MMC and OMC, making these measures redundant with those reported in seconds.

To evaluate alignment in the extraction of movement quality measures between the two systems, we visually assessed the measures from MMC and OMC using correlation plots. Additionally, Pearson correlations were calculated for each movement quality measure, both for individual trials ($r_s$) and for averaged movement measure ($r_{av}$) across all affected and unaffected trials per participant.

\section{Results}
\begin{figure}[!b]
    \centering
    \includegraphics[width=\columnwidth]{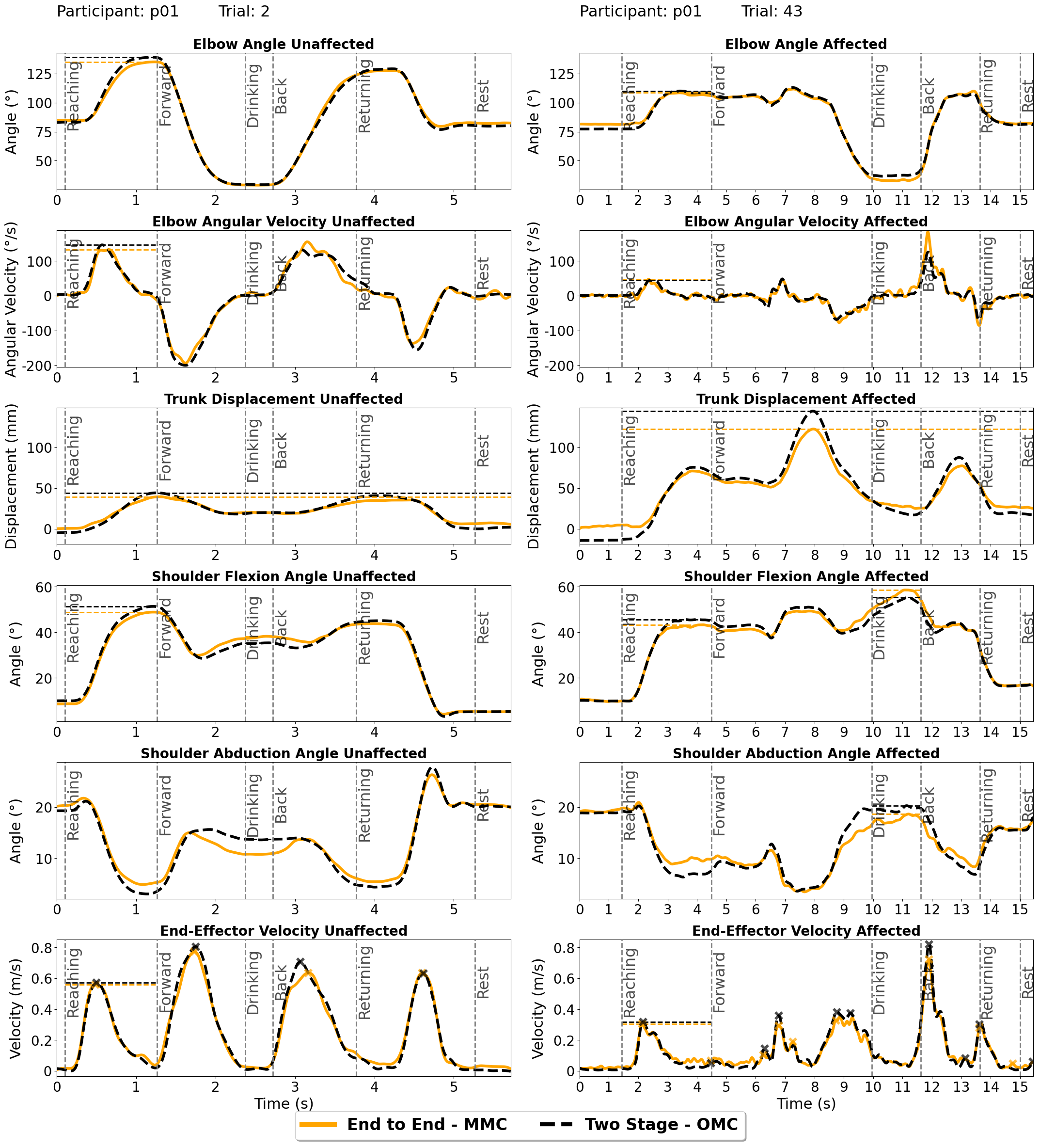}
    \caption{Exemplary kinematic trajectories of participant p01, showing one trial each for the affected and unaffected arm. Horizontal lines outline the movement quality measures, and the \texttt{x}'s on the end-effector velocity mark each movement unit.}
    \label{fig:Kinematics_examplel}
\end{figure}

\subsection{Study Population}
An ad-hoc sample of fifteen stroke patients (9 males, 6 females) was included in this study, with a mean age of 71 years (SD = 15) and an average time since stroke of 24 months. Participants had mild to moderate upper limb impairment, as indicated by a mean Fugl-Meyer Upper Limb Assessment score of 56.3 (SD = 8.4), with scores ranging from 39 to 66.

\subsection{Data and Processing}
In total 1160 trials of the drinking task of 15 participants were included with a median of 40 [39, 40] trials per affected arm and 39 [38, 41] trials per unaffected arm per participant. Trials were manually inspected, resulting in a 6\% exclusion rate. Of the excluded trials, 51 were due to incorrect movements, 13 due to end-to-end MMC reconstruction failures, and 5 due to synchronization issues between the two systems.
Calculating inverse kinematics in OpenSim with cleaned OMC data required approximately 30 hours on a 13th Gen Intel® Core™ i9-13900H processor (2.60 GHz) with 32 GB of RAM. The entire end-to-end MMC process took around 12 days to complete all trials, leveraging an NVIDIA RTX 3080 GPU. The majority of this time was spent extracting 2D keypoints from a total of 5800 videos (comprising 5 cameras across 1160 trials).

\subsection{System comparison}
\subsubsection{Kinematic Trajectories}
\begin{table}[t]
\centering
\renewcommand{\arraystretch}{1.2}  
\caption{Medians and [IQRs] for RMSE, $r$, bias, and time lag of kinematic trajectories, calculated across all patients and differentiating between the affected and unaffected arms.}
\begin{tabular}{@{} >{\raggedright\arraybackslash}p{1.5cm} @{\hskip 1pt} p{2cm} @{\hskip 1pt} p{2.6cm} @{\hskip 1pt} p{2.6cm} @{}}
\toprule
Kinematic & Value & Unaffected arm & Affected arm \\
\midrule
\multirow{4}{*}{\rotatebox{90}{\shortstack{End-Effector \\ Velocity}}} & r & 0.99 [0.97, 0.99] & 0.99 [0.96, 0.99] \\
 & RMSE (m/s) & 0.04 [0.03, 0.07] & 0.03 [0.03, 0.05] \\
 & Bias (m/s) & 0.00 [-0.01, 0.00] & 0.00 [-0.01, 0.00] \\
 & Time Lag (s) & 0.00 [0.00, 0.02] & 0.01 [0.00, 0.02] \\
\midrule
\multirow{4}{*}{\rotatebox{90}{\shortstack{Elbow \\ Angular \\ Velocity}}} & r & 0.96 [0.94, 0.97] & 0.96 [0.93, 0.97] \\
 & RMSE (°/s) & 16.11 [12.13, 20.95] & 13.12 [10.54, 17.54] \\
 & Bias (°/s) & 0.33 [-0.62, 1.22] & 0.26 [-0.58, 1.05] \\
 & Time Lag (s) & 0.02 [0.00, 0.03] & 0.02 [-0.02, 0.03] \\
\midrule
\multirow{4}{*}{\rotatebox{90}{\shortstack{Elbow \\ Extension}}} & r & 0.99 [0.98, 0.99] & 0.99 [0.98, 1.00] \\
 & RMSE (°) & 5.03 [3.83, 6.69] & 4.18 [3.16, 5.58] \\
 & Bias (°) & -7.52 [-13.84, -0.47] & -5.86 [-10.78, -2.45] \\
 & Time Lag (s) & 0.00 [-0.02, 0.03] & 0.02 [-0.02, 0.03] \\
\midrule
\multirow{4}{*}{\rotatebox{90}{\shortstack{Shoulder \\ Flexion}}} & r & 0.99 [0.99, 1.00] & 0.99 [0.99, 1.00] \\
 & RMSE (°) & 2.39 [1.86, 3.03] & 2.53 [1.73, 3.38] \\
 & Bias (°) & -23.55 [-26.57, -20.40] & -23.57 [-26.17, -19.45] \\
 & Time Lag (s) & -0.02 [-0.03, 0.02] & 0.00 [-0.03, 0.02] \\
\midrule
\multirow{4}{*}{\rotatebox{90}{\shortstack{Shoulder \\ Abduction}}} & r & 0.88 [0.78, 0.93] & 0.89 [0.83, 0.94] \\
 & RMSE (°) & 1.97 [1.37, 2.71] & 2.13 [1.50, 2.77] \\
 & Bias (°) & 7.42 [5.23, 11.11] & 7.47 [3.52, 11.65] \\
 & Time Lag (s) & 0.03 [0.00, 0.05] & 0.02 [-0.02, 0.03] \\
\midrule
\multirow{4}{*}{\rotatebox{90}{\shortstack{Trunk \\ Displacement}}} & r & 0.95 [0.91, 0.98] & 0.98 [0.91, 0.99] \\
 & RMSE (mm) & 5.41 [4.16, 7.00] & 6.18 [4.30, 10.71] \\
 & Bias (mm) & -4.62 [-8.10, -1.01] & -6.76 [-9.57, -3.39] \\
 & Time Lag (s) & -0.02 [-0.07, 0.02] & -0.02 [-0.05, 0.02] \\
\bottomrule
\end{tabular}
\label{tab:Kinematics_results}
\end{table}
Visual inspection of the kinematic trajectories demonstrated a very high level of agreement between OMC and MMC across all kinematic trajectories (see Fig. \ref{fig:Kinematics_examplel}).

Quantitative analysis further supported this (see Table \ref{tab:Kinematics_results}), with high correlation coefficients ($r$) observed between the two systems. Notably, shoulder flexion and elbow extension showed strong correlations ($r = 0.99$), indicating close alignment in these key movements. Among the kinematic trajectories, shoulder abduction on the unaffected side had the lowest correlation ($r = 0.88$ [$0.78$, $0.93$]), though the RMSE remained low at $1.97^\circ$ [$1.37^\circ$, $2.71^\circ$].

Overall, RMSE values were low, particularly for shoulder flexion and shoulder abduction (both approximately $2^\circ$). Elbow extension showed the highest RMSE among joint angle trajectories, with a median RMSE around $4^\circ$–$5^\circ$.

The trial-to-trial variability of bias in kinematic trajectories between OMC and MMC was low for each participant, as shown in Table \ref{tab:MeanIQR}, which presents the mean variability (IQR of bias per kinematic trajectory) across participants: $< 2^\circ$ for shoulder, $< 4^\circ$ for elbow extension, $< 2^\circ/\text{s}$ for elbow angular velocity, and around 3 mm for trunk displacement. However, absolute bias varied between participants, as reflected in the higher bias IQR in Table \ref{tab:Kinematics_results}, which shows the mean bias and IQR across participants.

Temporal alignment between systems remained consistent across all kinematic trajectories, with small discrepancies ($< 0.03$ s) primarily in shoulder and trunk metrics. 

\begin{table}[!t]
\centering
\caption{Mean IQR of bias across patients, representing the average variability of bias within a patient's trials.}
\begin{tabular}{llcc}
\toprule
Kinematic & bias & Unaffected arm & Affected arm \\
\midrule
End-Effector Velocity & $\overline{IQR}$ (m/s) & 0.01 & 0.01 \\
Elbow Angular Velocity & $\overline{IQR}$ (°/s) & 1.83 & 1.57 \\
Elbow Extension & $\overline{IQR}$ (°) & 3.69 & 3.88 \\
Shoulder Flexion & $\overline{IQR}$ (°) & 1.34 & 1.14 \\
Shoulder Abduction & $\overline{IQR}$ (°) & 1.20 & 0.91 \\
Trunk Displacement & $\overline{IQR}$ (mm) & 2.86 & 3.10 \\
\bottomrule
\end{tabular}
\label{tab:MeanIQR}
\end{table}

\subsubsection{Movement quality measures}
\begin{figure*}[!t]
    \centering
    \includegraphics[width=2\columnwidth]{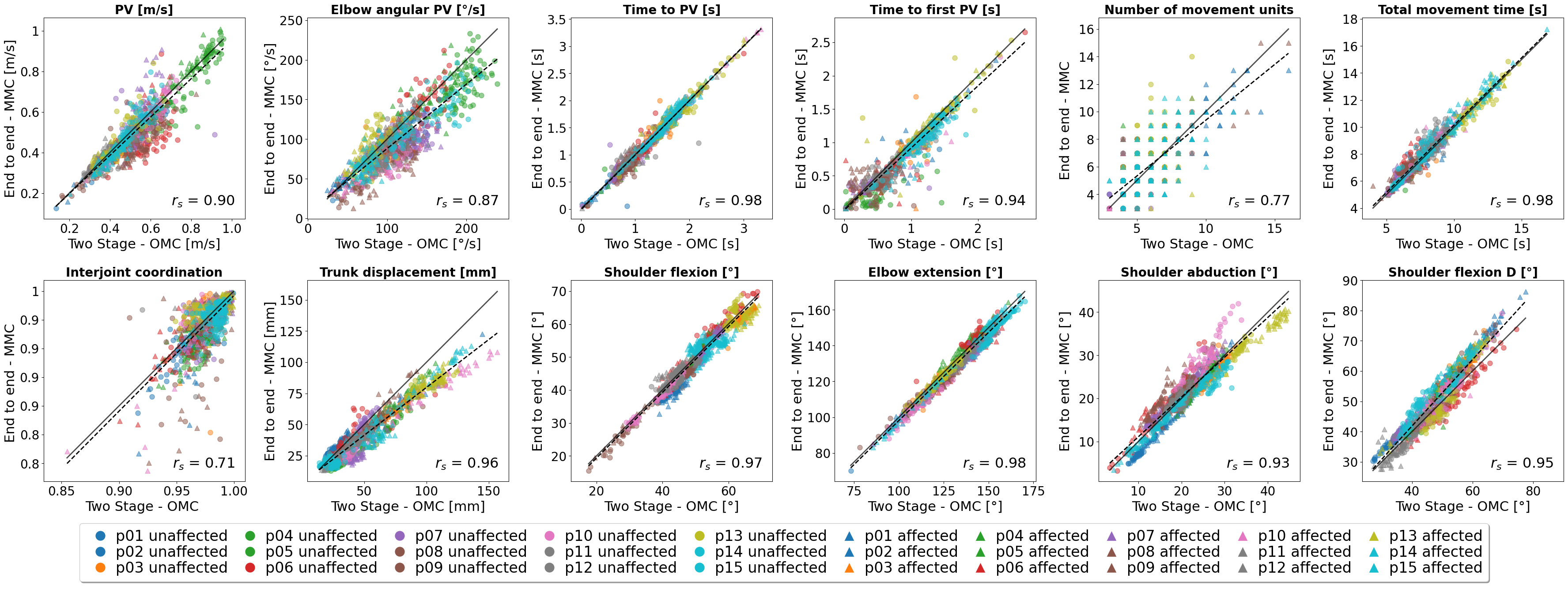}
    \caption{Correlation plot for each measure, comparing two-stage OMC and end-to-end MMC. Each data point represents a single trial. Trials are color-coded per participant and shape-coded per arm.}
    \label{fig:measure_plots}
\end{figure*}

We observed high correlations between systems for trial-by-trial comparisons, with improved correlations when averaging over trials ($ 0.91 < r_{av} < 1$, except for interjoint coordination, see Table \ref{tab:MurphyMeasures}).
Joint angle measures showed high correlations overall, with shoulder flexion ($r_s = 0.97$, $r_{av} = 0.98$), elbow extension ($r_s = 0.98$, $r_{av} = 0.99$), and shoulder abduction ($r_s = 0.93$, $r_{av} = 0.94$). Interjoint coordination had the lowest correlation ($r_s = r_{av} = 0.71$) with several outliers, while elbow angular PV showed lower correlation ($r_s = 0.87$, $r_{av} = 0.93$) and broader distribution of values (see Fig. \ref{fig:measure_plots}).
End-effector velocity measures exhibited high correlations (0.90 $< r_s <$ 0.98, 0.96 $< r_{av} <$ 1), except for the number of movement units ($r_s = 0.77$, $r_{av} = 0.91$) which also showed a broader distribution of values (see Fig. \ref{fig:measure_plots}).
Trunk displacement showed high correlation ($r_s = 0.96$, $r_{av} = 0.97$), though MMC systematically underestimated values compared to OMC at higher magnitudes (see Fig. \ref{fig:measure_plots}).

\section{Discussion}
This is the first study comparing biomechanics-based kinematics from OMC with the end-to-end MMC approach proposed by Cotton et al. \cite{cotton_differentiable_2024}. Few previous studies have compared biomechanically modeled upper limb kinematics from MMC to OMC using the commercial THEIA3D MMC (e.g., shadow boxing in able-bodied adults \cite{lahkar_accuracy_2022}, Box and Block in able-bodied children \cite{hansen_validation_2024}). To our best knowledge, this is also the first study to compare biomechanically modeled upper limb kinematics between OMC and MMC in neurological patients.

Overall, we find a high level of agreement between the systems for the kinematic trajectories of the drinking task in stroke patients and a strong correlation across the majority of movement quality measures.

The end-to-end MMC approach demonstrated high robustness, with failed reconstructions in only 1.6\% of trials (13 out of 1160).
For end-effector velocity, the small bias and low $\text{RMSE}$ indicate precise wrist keypoint detection and consistent scaling across methods, likely supported by bilevel optimization, enabling direct comparability of end-effector velocity between systems.

\begin{table}[!t]
\centering
\caption{Movement quality measures comparing OMC and end-to-end MMC: $r_s$ represents the correlation across all trials, while $r_{av}$ denotes the correlation for averaged participant trials per arm between the two systems.}
\begin{tabular}{lcc}
\toprule
Movement Quality Measure & $r_s$ & $r_{av}$ \\
\midrule
PV [m/s] & 0.90 & 0.96 \\
Elbow angular PV [°/s] & 0.87 & 0.93 \\
Time to PV [s] & 0.98 & 1.00 \\
Time to first PV [s] & 0.94 & 0.99 \\
Number of movement units [n] & 0.77 & 0.91 \\
Total movement time [s] & 0.98 & 1.00 \\
Interjoint coordination & 0.71 & 0.71 \\
Trunk displacement [mm] & 0.96 & 0.97 \\
Shoulder flexion [°] & 0.97 & 0.98 \\
Elbow extension [°] & 0.98 & 0.99 \\
Shoulder abduction [°] & 0.93 & 0.94 \\
Shoulder flexion D [°] & 0.95 & 0.97 \\
\bottomrule
\end{tabular}
\label{tab:MurphyMeasures}
\end{table}

Differences in elbow angle (median $\text{RMSE}$ around $5^\circ$), trunk displacement (median $\text{RMSE}$ around $6 \text{mm}$), and shoulder angles (median $\text{RMSE}$ around $2^\circ$) are small and remaining differences between OMC and MMC may reflect the influence of marker placement and model variations, which can cause slight discrepancies in trajectories. 
The differences in kinematic trajectories between OMC and MMC fall within the session-to-session variability observed with OMC alone \cite{uchida_conclusion_2022}, suggesting that the end-to-end MMC is nearing the accuracy and precision of OMC. Our results are consistent with, and slightly better than, prior MMC-OMC comparisons using the commercial THEIA3D system, which reported joint angle differences with RMSE $< 6^\circ$ between systems \cite{hansen_validation_2024}. Comparisons of MMC and OMC in upper limb kinematics remain limited, and no studies have yet evaluated non-commercial, biomechanics-based MMC kinematics against OMC. The end-to-end MMC approach presented here demonstrates an accuracy comparable to THEIA3D while outperforming earlier custom MMC systems tested for the drinking task \cite{huber_computer_2024}. MMC shows promise as a reliable alternative to OMC, offering significantly reduced setup and manual processing times.

Temporal discrepancies between the systems were negligible, with time lags ranging from $0$ to $0.03 \, \text{s}$, supporting the reliability of the end-to-end MMC approach in capturing temporal aspects of movement with high accuracy.

Overall, the correlation between OMC and end-to-end MMC is high across most movement quality measures, with few outlier trials observed. Averaging trials generally improves correlation (e.g., $r_{av} = 1.00$ for time to peak velocity and total movement time), aligning with recommended clinical practice for performance stability \cite{frykberg_how_2021}. While the number of trials needed to ensure performance stability is established for OMC \cite{frykberg_how_2021}, slight adjustments may be needed for MMC to account for measurement uncertainty. Averaging also aids in outlier detection, valuable for robustness in clinical applications.

Movement quality measures based on joint angles, such as shoulder flexion ($r_s = 0.97$, $r_{av} = 0.98$) and elbow extension ($r_s = 0.98$, $r_{av} = 0.99$), as well as end-effector velocity measures like peak velocity and time to peak velocity, show strong agreement between systems, indicating high accuracy in detecting peak amplitudes in these key kinematic trajectories. Slight underestimation in trunk displacement (see Fig. \ref{fig:measure_plots}) may result from model differences in chest marker placement, suggesting potential for model refinement.

Metrics like the number of movement units ($r_s = 0.77$, $r_{av} = 0.91$) and interjoint coordination ($r_s = r_{av} = 0.71$) show lower correlation, where small kinematic trajectory variations can produce larger discrepancies. Replacing the number of movement units with alternative smoothness metrics, such as Log Dimensionless Jerk, could improve the reliability of smoothness assessment.

Observed trial-to-trial biases in kinematic trajectories between OMC and MMC were consistent across trials within sessions, as evidenced by small IQR (approximately $1^\circ-3^\circ$ for joint angles, 0.01 m/s for end-effector velocity and 3 mm for trunk displacement). This suggests a systematic bias per session, likely due to differences in marker/keypoint locations between systems (OMC anatomical landmarks vs. MMC-detected keypoints) or biomechanical model variations.
The session-to-session (patient-to-patient) variation in bias may stem from variability in both OMC and MMC systems. In MMC, keypoint detection may slightly differ between individuals, while in OMC, minor differences in marker placement and registration can significantly impact inverse kinematics. Uchida and Seth \cite{uchida_conclusion_2022} for example demonstrated that a 1–2 cm marker registration uncertainty can lead to a 5°–10° difference in knee flexion angles. Additionally, a prior study using the commercial THEIA3D MMC system demonstrated lower inter-session variability in MMC compared to OMC \cite{kanko_inter-session_2021}. Thus, we assume that session-to-session bias variability is primarily due to OMC variability. However, definitive conclusions require further data, such as multi-session recordings of the same participants or true ground-truth bone movement data from dual-plane fluoroscopy.

Although the findings are promising, this study has several limitations. First, while OMC is generally considered the gold standard, it is not an absolute ground truth due to potential errors from marker misplacement and soft-tissue artifacts. Additionally, the used upper limb biomechanical models lack certain degrees of freedom (e.g., finger motion), which could provide a more complete assessment. For example, \cite{firouzabadi_biomechanical_2024} extends the end-to-end MMC optimization approach to a full arm and hand model.
The models in OpenSim for OMC and MuJoCo for end-to-end MMC are not identical, which may impact direct data comparability between the systems. Furthermore, while OMC uses a two-stage process, MMC employs an end-to-end approach; using OMC data in an end-to-end model could help better isolate the effects of marker-based versus markerless capture.

Phase classification was conducted exclusively with OMC data, meaning that MMC phase classification still requires validation for standalone application. Lastly, the study focuses on a specific patient population with mild to moderate impairment performing a single task. Results may vary with other tasks or populations, potentially introducing additional challenges and considerations for broader application.

Future research should focus on collecting datasets with simultaneous biplanar fluoroscopy, MMC, and OMC recordings to establish true ground truth for contextualizing clinical measurement uncertainty as in \cite{unger_upper_2024}. Additionally, research should investigate the impact of camera number and recording angles on MMC accuracy (including monocular setups) and expand the analysis to encompass all upper body degrees of freedom, including fingers \cite{firouzabadi_biomechanical_2024}.

\section{Conclusion}
MMC accuracy and precision continue to improve thanks to better keypoint detectors and sophisticated GPU-accelerated biomechanical modeling, reaching levels comparable to OMC, as demonstrated in this study. MMC eliminates the burden of marker placement and significantly reduces costs and application complexity, requiring only a minimal setup with consumer-grade cameras. Our results show promise for clinical application, suggesting that MMC could enable the collection of accurate, precise movement data in routine clinical settings, providing a valuable tool for novel, more precise biomarkers in movement rehabilitation.

\section*{Acknowledgment}
We sincerely thank the participants of this study.


\bibliographystyle{ieeetr}
\bibliography{references}

\end{document}